# Can Large Language Models Make the Grade?

An Empirical Study Evaluating LLMs Ability to Mark Short Answer Questions in K-12 Education

Owen Henkel[†]
Department of Education
Univeristy of Oxford
owen@education.ox.ac.uk

Adam Boxer
Carousel Learning

Libby Hills | Bill Roberts
Jacobs Foundation | Legible Labs

## ABSTRACT

This paper presents reports on a series of experiments with a novel dataset evaluating how well Large Language Models (LLMs) can mark (i.e. grade) open text responses to short answer questions, Specifically, we explore how well different combinations of GPT version and prompt engineering strategies performed at marking real student answers to short answer across different domain areas (Science and History) and grade-levels (spanning ages 5-16) using a new, never-used-before dataset from Carousel, a quizzing platform. We found that GPT-4, with basic few-shot prompting performed well (Kappa, 0.70) and, importantly, very close to human-level performance (0.75). This research builds on prior findings that GPT-4 could reliably score short answer reading comprehension questions at a performance-level very close to that of expert human raters. The proximity to human-level performance, across a variety of subjects and grade levels suggests that LLMs could be a valuable tool for supporting low-stakes formative assessment tasks in K-12 education and has important implications for real-world education delivery.

## 1 Introduction

Assessment and feedback are crucial components of the learning process but are resource intensive to do regularly at scale [4, 21]. Unlike the high-pressure environment of traditional high stakes examinations, formative assessments are generally intended to be diagnostic, enabling students and teachers to adapt their approach within or in-between lessons to maximize learning. Also sometimes referred to as assessment for learning, formative assessment has been shown to lead to significant improvements in learning outcomes [12], however, scaling formative assessment practices have traditionally proven to be challenging due to the significant costs and logistical demands involved [4].

Closed-response assessment questions, such as multiple-choice and true/false, are commonly used in formative assessment, and have the benefit of being efficient to grade [18]. However, they have several drawbacks, such as the possibility of students relying on test-taking strategies, a potential lack of face validity, and the complexity in generating multiple answer options [2, 16]. In contrast, open-ended and short answer questions require the student to answer a question using their own words in a few sentences [18]. Many researchers argue that they decrease the influence of test-taking strategies, have greater face validity, have lower risk of floor effects, and may be better suited to evaluate certain subprocesses of the skill being assessed [3, 8]. and for many formative assessment tasks may be preferable [4, 21]. However, the process of grading open-ended questions can be resource-intensive and expensive, which limits their widespread use [15].

Automatic short answer grading (ASAG) has been an active area of research for nearly a decade, but it has historically been challenging to do easily and effectively enough, for widespread use in educational settings [7]. Up recently, most SOTA approaches relied primarily on transfer learning and fine-tuning models for specific tasks [10], which models required extensive technical expertise and large datasets [17, 20]. These models often struggled with domain shift when implemented in educational contexts [5, 19]. Mitigating domain shift can be particularly challenging because even tasks that appear to be similar in their may have subtle differences that are not immediately evident but can greatly impact the model's performance [9, 22].

A notable shift has occurred in the paradigm underlying the most recent generation of Large Language Models (LLMs), exemplified by models like GPT-4, Bard, and Claude. These frontier generative models are also better able to generalize across different tasks and require significantly less technical expertise to implement. There is growing evidence that these models can complete evaluation tasks on novel datasets with only minimal prompt engineering [11, 14]. If true, this would greatly enhance their usability and potential for use in real world educational environments. However, despite the enthusiasm surrounding the potential of LLMs. Indeed, prior work has suggested that LLMs do have promise for acting as formative assessment tool, and were found perform equivalently to expert human raters, when evaluating short answer reading comprehension questions [13]. However, little is known about the subjects in which they perform best, the grade levels they are most effective with, and what types of responses they can accurately assess, and whether they bound of model performance as task complexity increases. However, if LLMs can accurately mark open-ended formative assessment times, the time savings for teachers would be substantial, and could facilitate more frequent and effective formative assessment.

To explore these questions, we first built a new dataset of 1,700 marked student responses to open-ended History and Science questions using data from an online quizzing platform called Carousel. We then used various configurations of GPT-4 to classify student responses from this dataset, allowing for a comparison between human and LLM performance. Our preliminary findings suggest promise for the potential of LLMs to reliably evaluate a range of formative assessment tasks, beyond foundational literacy. We found that the optimal model configuration (GPT-4 with few-shot prompting) produced close to human-level agreement (Kappa, 0.70 versus 0.75) across different grades and subjects.

## 2 Current Study

### 2.1 Carousel Short Answer Dataset

Carousel is a UK-based online learning platform that has developed an online quizzing tool for students aged 5-18 that utilizes open-ended short-answer questions. Teachers upload question banks to create quizzes on the platform, and students respond to these questions and self-assess their answers, which are then moderated by teachers. Carousel has 135,000 active student users, 8,200 active teacher users, and over 100,000,000 questions answered by students. We have partnered with Carousel to create a new dataset consisting of 1710 student responses to 12 open-response questions, across two domain areas (Science and History) and three different educational stages (Key Stage 2, Key Stage 3, and Key Stage 4).

For each subject and grade-level two questions were then selected (one easy and one hard) to also explore model performance across different question types. Questions were categorized as easy or hard by the Carousel team based on the level of item complexity contained in the question and level of abstraction. Each item in the dataset includes a question, an exemplar "correct" answer, a student response, and a rating ('correct' or 'incorrect') from an expert marker, all of whom qualified teachers were.

Once this data had been compiled it was pre-processed to remove any null answers (where a student left it blank or indicated they were unable to answer the question), and 'perfect' answers from students. Perfect answers were answers that exactly matched the exemplar answer given by Carousel. This step ensured that student responses in the final dataset were ones where there was a degree of ambiguity as to whether the responses were correct or not, therefore requiring an element of judgment to be made to assess its correctness rather than just matching the response to the exemplar answer.

### 2.2 Human Evaluation of Student Answers

Carousel leveraged its large community of users and teachers to help evaluate student answers. After putting out a call for volunteers through their network, Carousel found 38 volunteer markers, who were current teachers of the subjects they were evaluating (Science and History). When the two raters disagreed (approx. 13% of the time), a third expert rater was brought in to cast a tie-breaking vote, which was used as the ground truth for whether a student answer was correct or incorrect.

Table 1 summarizes the percentage of student responses that were judged to be correct according to the ground truth answer, as well as the interrater agreement. At an aggregate level, students were judged to answer the questions correctly 53% of the time. It is important to note that 50% accuracy does not imply random guessing, as the accuracy rate expected by guessing or chance alone would be approximately 0% for open-response questions.

**Table 1: Human Evaluation of Student Answers**

| | |
|---|---|
| Total Questions | 1710 |
| Correct Student Answer Rate | 53% |
| Human Rater Agreement, Cohen's Kappa | 0.75 |
| Correlation between subject (history) and correctness, $\varphi^2$ | 0.255 |
| Correlation between difficulty (easy) and correctness, $\varphi^2$ | 0.198 |
| Correlation between key stage and correctness, Cramer's V | 0.207 |

Looking at whether the subject, question difficulty, and key stage impacted the rate at which students answer questions correctly, we found modest correlations between all three. This result is not particularly surprising, and there are several plausible explanations for this, including that the specific elements of a question were a bigger determinant of question difficulty rather than subject and/or key stage, desirable self-sorting (i.e., students working on questions at an appropriate level of difficulty), and heterogeneity between students (it wasn't the same set of students answering all questions).

### 2.3 Interrater Agreement

Table 2 reports inter-rater agreement (i.e., before any disagreements were resolved by tie-breaking votes) using both percentage agreement and Cohen's Kappa. The human raters, qualified teachers, agreed with each other on whether the student was correct or incorrect 87% of the time. Because raters were given a binary classification task (correct/incorrect), the rate of agreement expected by chance would be 50%. We use Cohen's Kappa as the preferred chance-adjusted measure of interrater reliability and find a K value of 0.75, which is considered medium-high [1].

Looking at whether the subject (Science vs. History), question difficulty, and key stage impacted the rate at which human raters agreed, we found only small correlations. This suggests that the rater agreement was consistent across these different categories. This result is relatively intuitive, as teachers graded items from the subjects and ages that they taught.

**Table 2: Interrater Agreement**

| | |
|---|---|
| Total Questions | 1710 |
| Human Rater Agreement, Precent | 87% |
| Human Rater Agreement, Cohen's Kappa | 0.75 |
| Correlation between subject (history) and agreement, φ² | 0.017 |
| Correlation between difficult (easy) and agreement, φ² | 0.171 |
| Correlation between key stage and agreement, Cramer's V | 0.173 |

Qualitatively, rater disagreement typically occurred when a student answer contained some of the correct information but either omitted some important information or added extraneous or inaccurate information. This is intuitive, as these are precisely the "edge cases" where individual teachers' judgment might differ. Relatively straightforward tasks with more objective criteria are likely to have high agreement among individuals, while tasks that depend more on expertise or are inherently more ambiguous almost always have lower agreement level.

## 2.4 Model Evaluation of Student Answers

The goal of our study was to evaluate students' short answers to open-ended questions in Science and History. For each question, the model was presented with the question and candidate answer and tasked with predicting whether the candidate answer was correct or not. These model predictions were then compared to the ground truth labels generated by expert human raters. To evaluate the impact of employing generative Large Language Models (LLMs) in comparison to human raters, we presented the model with the same task as the human raters. Given the topic, question, correct answer, and candidate answer, the model was asked to predict whether the candidate answer was correct or not. These model predictions were then compared to the ground truth labels.

We used a simple prompt and developed a Python script that (a) retrieves the pertinent question and exemplar answer from a data table, (b) places them at the appropriate place in the prompt, (c) makes a call containing the prompt to the OpenAI API, and (d) saves the API output. We further experimented with two different versions of GPT: gpt-3.5-turbo-0125 and gpt-4-0125-preview, as well as using few-shot examples of student responses. We consistently set the model temperature to zero to reduce output variability. The four variations are summarized in Table 3 below.

## 2.4 Results

Consistent with both the authors' prior work and that of others, the few-shot GPT-4 model configuration achieved the highest performance. While few-shot prompting modestly improved performance, using GPT-4 vs. GPT-3.5 dramatically improved performance. The top-performing combination, GPT-4 using few-shot prompting, had a Cohen's kappa score of 0.70.

**Table 3: Model Evaluation of Student Answers**

| | GPT 3.5 Zero-Shot | GPT 3.5 Few-Shot | GPT 4 Zero-Shot | GPT 4 Few-Shot |
|---|---|---|---|---|
| Correctly Predicted | 71 % | 72 % | 85 % | 84% |
| Precision | 0.94 | 0.93 | 0.85 | 0.87 |
| Recall | 0.50 | 0.51 | 0.85 | 0.85 |
| F1 | 0.65 | 0.66 | 0.85 | 0.86 |
| Percent Agreement | 0.71 | 0.72 | 0.84 | 0.85 |
| Cohen's Kappa | 0.44 | 0.45 | 0.68 | 0.70 |

In line with our prior research [x], we found that GPT-4 performed substantially better than GPT-3.5. As can be seen in Table 3, this improvement was primarily due to GPT-4 dramatically improving recall while only slightly decreasing precision, specifically by ruling fewer student answers as being incorrect. Analogically, we might be able to understand this as GPT-3.5 being a "harsh-grader," classifying too many student answers as being incorrect.

Also, in line with prior research [6], few-shot learning modestly improved model performance, both with GPT-3.5 and GPT-4.0. The top-performing model, GPT-4 using few-shot, achieved a Cohen's kappa of 0.70, which was close to human-level performance (Kappa, 0.75). In a task such as evaluating open-response questions, expert human-rater agreement is likely to effectively be the ceiling of performance. In Table 4, we examine how various attributes of the question impacted the model's accuracy. We find only very small correlations between the subject, question difficulty, key stage, and the rate at which the model accurately graded the student responses. Interestingly, we do find a small to moderate correlation between incorrect student answers and model accuracy.

**Table 4: Correlation Between Model and Other Factors**

| | |
|---|---|
| Total Questions | 1710 |
| Model Agreement with Ground Truth, Cohen's Kappa | 0.70 |
| Student Answer type (incorrect) and model accuracy, φ² | -0.28 |
| Correlation between difficult (easy) and model accuracy φ² | 0.10 |
| Correlation between subject (science) and rater agreement, φ² | 0.14 |
| Correlation between key stage and model accuracy, Cramer's V | 0.08 |

## 3 Discussion

In this experiment, we aimed to explore how well GPT-4 performs at marking open text responses to short answer questions and how performance varies across domain areas, grade levels, and question difficulty. Overall, we found that GPT-4 performed well human-level performance (Kappa, 0.75). That GPT-4 performs in line with expert human raters is consistent with prior work which tested model performance at scoring short-answer reading comprehension questions [13]. We found only very modest variation in model performance based on subject, grade-level, and question difficulty.

Our results from this experiment with Carousel build on our prior work focused on reading comprehension and, importantly, suggest that LLMs like GPT-4 could potentially be used for a variety of low-stakes assessment tasks across different domain areas and grades. While the extent to which we could test model performance across these question attributes was limited, and results should therefore be interpreted with caution, that model performance was relatively consistent across both subjects, grade levels, and question difficulty. is interesting and suggests potential applicability in multiple real-world settings.

Our findings build on previous research where we tested how reliably LLMs evaluated responses of 4th graders to open-response reading comprehension questions. In that experiment the top model performance (Kappa 0.950) was in line with expert human raters (Kappa 0.947), with GPT-4 being the best performer, as in this experiment. The gap between the model and humans was approximately equivalent. The lower overall agreement in this experiment with Carousel is likely because, in a task such as evaluating open-response questions, expert human-rater agreement is likely to highly correlate with model-human performance and be the ceiling of performance. As earlier discussed, there is a certain degree of judgment and interpretation when evaluating edge cases, and disagreement often represents inherent ambiguity in the task (i.e., the task isn't well-specified enough). For example, human raters may have based their choice of whether an answer was correct on their views regarding the importance of students using the correct spelling of key terms. This inherent ambiguity, which humans struggle with, means there is effectively a ceiling on the agreement you could expect from a model.

In this experiment, we also tested different configurations of models and, again as in our prior work, found that GPT-4 had the strongest performance. Similar to our prior work, in this experiment, we found that few-shot prompting very slightly improved model performance (Kappa 0.70 with few-shot prompting versus Kappa 0.68 with zero-shot prompting). While this difference is not substantial, it could be explained due to the lack of context that we provided in this experiment versus our previous experiment. As we did not provide context alongside the question-and-answer pairs, examples may have played a greater role in improving model performance.

The proximity to human-level performance we found in this experiment suggests that GPT-4 could be useful for low-stakes assessment tasks in K-12 education. This is especially the case when the relative timesaving for teachers is considered. Our rough conservative estimate, assuming the use of a platform such as Carousel rather than natural marking, suggests that it would have taken teachers at least 11 hours to mark the number of questions (1,700) included in this dataset. Using GPT-4 to mark the same dataset, after setting up a generic script, took approximately 2 hours, clearly a significant timesaving for a similar level of accuracy.

Further work could also explore how alternative categorization for different types of student responses could be more predictive of model performance. The categories we looked at here were based on how Carousel structures their quizzes and question banks, which reflects the organization of UK schools and curriculum by subject and grade level. However, it could be that other features of text are more predictive of model performance than domain area and grade level. Exploring this would require further research, but one hypothesis is that the degree of judgment required to assess the correctness of a response (in other words, how ambiguous the task is) could be one such feature.

## 4 Conclusion

This study reports on a series of experiments with a novel dataset evaluating LLMs can mark open text responses to short answer questions, across various by subject areas and grade levels. We found that GPT-4 performance (with minimal prompt engineering) was in line with the performance of expert human raters. Model performance was consistent with human performance, itself a valuable finding, but this outcome was achieved significantly more efficiently than if marking had taken place manually. When taken together with prior work [13], which found that GPT-4 performed well at assessing literacy tasks, this experiment suggests that LLMs could become a useful formative assessment tool.

Moving forward, further research should focus on expanding the scope of the datasets used to evaluate model performance, including the development of novel, non-standard questions that are unlikely to have been included in the model's training data. Additionally, exploring the factors that influence model performance, such as the degree of judgment required to assess the correctness of a response, could provide valuable insights into how LLMs can be optimized for use in educational settings.

Our findings contribute to the growing body of research on the application of LLMs in educational settings. As the capabilities of these models continue to evolve, it is crucial for researchers and educators to collaborate in exploring their potential benefits and limitations. By conducting empirical studies using real-world student data, we can gain a better understanding of how LLMs can be effectively integrated into educational practices to support student learning and assessment.

## ACKNOWLEDGMENTS

We would like to than the teacher partners at Carousel for their work in evaluating student responses.

## REFERENCES

[1] Banerjee, M., Capozzoli, M., Mcsweeney, L. and Sinha, D. 2008. Beyond Kappa: A Review of Interrater Agreement Measures. *Canadian Journal of Statistics*. 27, (Dec. 2008), 3–23. DOI:https://doi.org/10.2307/3315487.

[2] Bellinger, J.M. and DiPerna, J.C. 2011. Is fluency-based story retell a good indicator of reading comprehension? *Psychology in the Schools*. 48, 4 (Apr. 2011), 416–426. DOI:https://doi.org/10.1002/pits.20563.

[3] van den Bergh, H. 1990. On the Construct Validity of Multiple- Choice Items for Reading Comprehension. *Applied Psychological Measurement*. 14, 1 (Mar. 1990), 1–12. DOI:https://doi.org/10.1177/014662169001400101.

[4] Black, P. and Wiliam, D. 2009. Developing the theory of formative assessment. *Educational Assessment, Evaluation and Accountability*. 21, 1 (2009), 5–31. DOI:https://doi.org/10.1007/s11092-008-9068-5.

[5] Bommasani, R. et al. 2022. On the Opportunities and Risks of Foundation Models. arXiv.

[6] Brown, T.B., Mann, B., Ryder, N., Subbiah, M., Kaplan, J., Dhariwal, P., Neelakantan, A., Shyam, P., Sastry, G., Askell, A., Agarwal, S., Herbert-

Voss, A., Krueger, G. and Henighan, T. 2020. Language Models are Few-Shot Learners. (2020).

[7] Burrows, S., Gurevych, I. and Stein, B. 2015. The Eras and Trends of Automatic Short Answer Grading. *International Journal of Artificial Intelligence in Education*. 25, 1 (2015), 60–117. DOI:https://doi.org/10.1007/s40593-014-0026-8.

[8] Cain, K. and Oakhill, J. 2007. *Children's comprehension problems in oral and written language a cognitive perspective*. Guilford Press.

[9] Elsahar, H. and Gallé, M. 2019. To Annotate or Not? Predicting Performance Drop under Domain Shift. *Proceedings of the 2019 Conference on Empirical Methods in Natural Language Processing and the 9th International Joint Conference on Natural Language Processing (EMNLP-IJCNLP)* (Hong Kong, China, 2019), 2163–2173.

[10] Fernandez, N., Ghosh, A., Liu, N., Wang, Z., Choffin, B., Baraniuk, R. and Lan, A. 2023. Automated Scoring for Reading Comprehension via In-context BERT Tuning. arXiv.

[11] Gilardi, F., Alizadeh, M. and Kubli, M. 2023. ChatGPT outperforms crowd workers for text-annotation tasks. *Proceedings of the National Academy of Sciences*. 120, 30 (Jul. 2023), e2305016120. DOI:https://doi.org/10.1073/pnas.2305016120.

[12] Hattie, J. 2010. *Visible learning: a synthesis of over 800 meta-analyses relating to achievement*. Routledge.

[13] Henkel, O., Hills, L., Roberts, B. and McGrane, J. 2023. Supporting Foundational Literacy Assessment in LMICs: Can LLMs Grade Short-answer Reading Comprehension Questions? (2023).

[14] Kuzman, T., Mozetič, I. and Ljubešić, N. 2023. ChatGPT: Beginning of an End of Manual Linguistic Data Annotation? Use Case of Automatic Genre Identification. arXiv.

[15] Magliano, J.P. and Graesser, A.C. 2012. Computer-based assessment of student-constructed responses. *Behavior Research Methods*. 44, 3 (2012), 608–621. DOI:https://doi.org/10.3758/s13428-012-0211-3.

[16] Magliano, J.P. and Millis, K.K. 2003. Assessing Reading Skill With a Think-Aloud Procedure and Latent Semantic Analysis. *Cognition and Instruction*. 21, 3 (2003), 251–283. DOI:https://doi.org/10.1207/S1532690XCI2103_02.

[17] Mayfield, E. and Black, A.W. 2020. Should You Fine-Tune BERT for Automated Essay Scoring? *Proceedings of the Fifteenth Workshop on Innovative Use of NLP for Building Educational Applications* (Seattle, WA, USA → Online, 2020), 151–162.

[18] Pearson, P.D. and Hamm, D.N. 2006. The Assessment of Reading Comprehension: A Review of Practices— Past, Present, and Future. *Children's reading comprehension and assessment*. Lawrence Erlbaum Associates.

[19] Perez, E., Kiela, D. and Cho, K. 2021. True Few-Shot Learning with Language Models. (2021).

[20] Pulman, S.G. and Sukkarieh, J.Z. 2005. Automatic short answer marking. *Proceedings of the second workshop on Building Educational Applications Using NLP - EdAppsNLP 05* (Ann Arbor, Michigan, 2005), 9–16.

[21] Shute, V.J. 2008. Focus on Formative Feedback. *Review of Educational Research*. 78, 1 (Mar. 2008), 153–189. DOI:https://doi.org/10.3102/0034654307313795.

[22] Zhao, S., Li, B., Reed, C., Xu, P. and Keutzer, K. 2020. Multi-source Domain Adaptation in the Deep Learning Era: A Systematic Survey. arXiv.